\documentclass{article}

\usepackage[english]{babel}
\usepackage{tikz}

\usepackage{adjustbox}
\usepackage{rotating}
\usepackage{subcaption}

\usetikzlibrary{shapes, arrows, positioning}
\usepackage[backend=biber, style=numeric, sorting=ynt]{biblatex}
\usepackage{pgfplots}
\usepackage{listings}
\usepackage{xcolor}
\usepackage{float}
\usepackage{csquotes}
\usepackage{tabularx}
\usepackage{booktabs}  
\usepackage{caption}
\pgfplotsset{compat=1.18}
\usepackage{fancyhdr}
\usepackage[letterpaper,top=2cm,bottom=2cm,left=3cm,right=3cm,marginparwidth=1.75cm]{geometry}

\usepackage{amsmath}
\usepackage{graphicx}
\usepackage[colorlinks=true, allcolors=blue]{hyperref}
\title{\textbf{On Generalization in Agentic Tool Calling: CoreThink Agentic Reasoner and MAVEN Dataset}}

\author{
Omkar Ghugarkar\textsuperscript{1} \quad
Vishvesh Bhat\textsuperscript{1,2} \quad
Julian McAuley\textsuperscript{2} \\
\\[-0.7em]
\textsuperscript{1}\textit{CoreThink AI} \\
\textsuperscript{2}\textit{University of California, San Diego}
}

\date{October 26, 2025}

\begin{document}
\maketitle

\thispagestyle{fancy}

\begin{abstract}
Generalization across Agentic tool-calling environments remains a key unsolved challenge in developing reliable agentic reasoning systems. While large language models (LLMs) demonstrate strong performance on isolated benchmarks, their ability to transfer reasoning strategies and coordinate tools across diverse domains is poorly understood. In this work, we conduct a large-scale evaluation of state-of-the-art LLMs on multiple tool-calling benchmarks—\textbf{BFCL v3}, \textbf{TauBench}, \textbf{Tau2Bench}, and \textbf{AceBench}—and introduce \textbf{MAVEN} (Math \& Physics Adversarial Verification \& Evaluation Network), a new out of distribution (OOD) benchmark designed to stress-test multi-step reasoning through explicit verification and adversarial task composition. Our results show that most current models achieve below 50\% accuracy on MAVEN, revealing a significant generalization gap across tool-use settings.

To address this, we present the \textbf{CoreThink Agentic Reasoner}, a framework that augments LLMs with a lightweight symbolic reasoning layer for structured decomposition and adaptive tool orchestration. Without additional training, it generalizes across all benchmarks, achieving state-of-the-art performance with 5--30\% improvements over existing baselines at roughly one-tenth the computational cost.
\end{abstract}

\section{Introduction}

The growing capabilities of large language models (LLMs) have enabled the development of autonomous, ``agentic'' systems that perform complex, multi-step tasks by planning, reasoning, and coordinating specialized tools. However, many of the current agentic systems often might struggle with reliably decomposing long-horizon tasks, verifying intermediate results, and generalizing to out-of-distribution challenges. Evaluating these capabilities requires robust benchmarks that stress-test multi-step reasoning and tool orchestration; datasets such as BFCL v3, Taubench, Tau2bench, and Acebench\cite{bfcl_v3}\cite{tau_bench}\cite{tau2_bench}\cite{acebench} have emerged as key metrics, while MAVEN (Math \& Physics Adversarial Verification \& Evaluation Network) pushes the boundary further by probing reasoning under adversarial, long-horizon conditions.

Despite the existence of these benchmarks, an over-reliance on a fixed set of evaluations raises critical questions about the true generalization capabilities of frontier models. High performance on some standard datasets does not necessarily indicate robust reasoning; models may simply exploit dataset-specific patterns or artifacts, a phenomenon we refer to as ``benchmark brittleness.''\cite{lunardi2025robustnessreliabilitybenchmarkbasedevaluation} This limitation is particularly concerning for real-world deployment, where agents must reliably tackle novel, unpredictable tasks. Compounding this challenge, the substantial computational cost of training and running these state-of-the-art models creates barriers for widespread experimentation.

\begin{figure}
\centering
\includegraphics[width=0.9\linewidth]{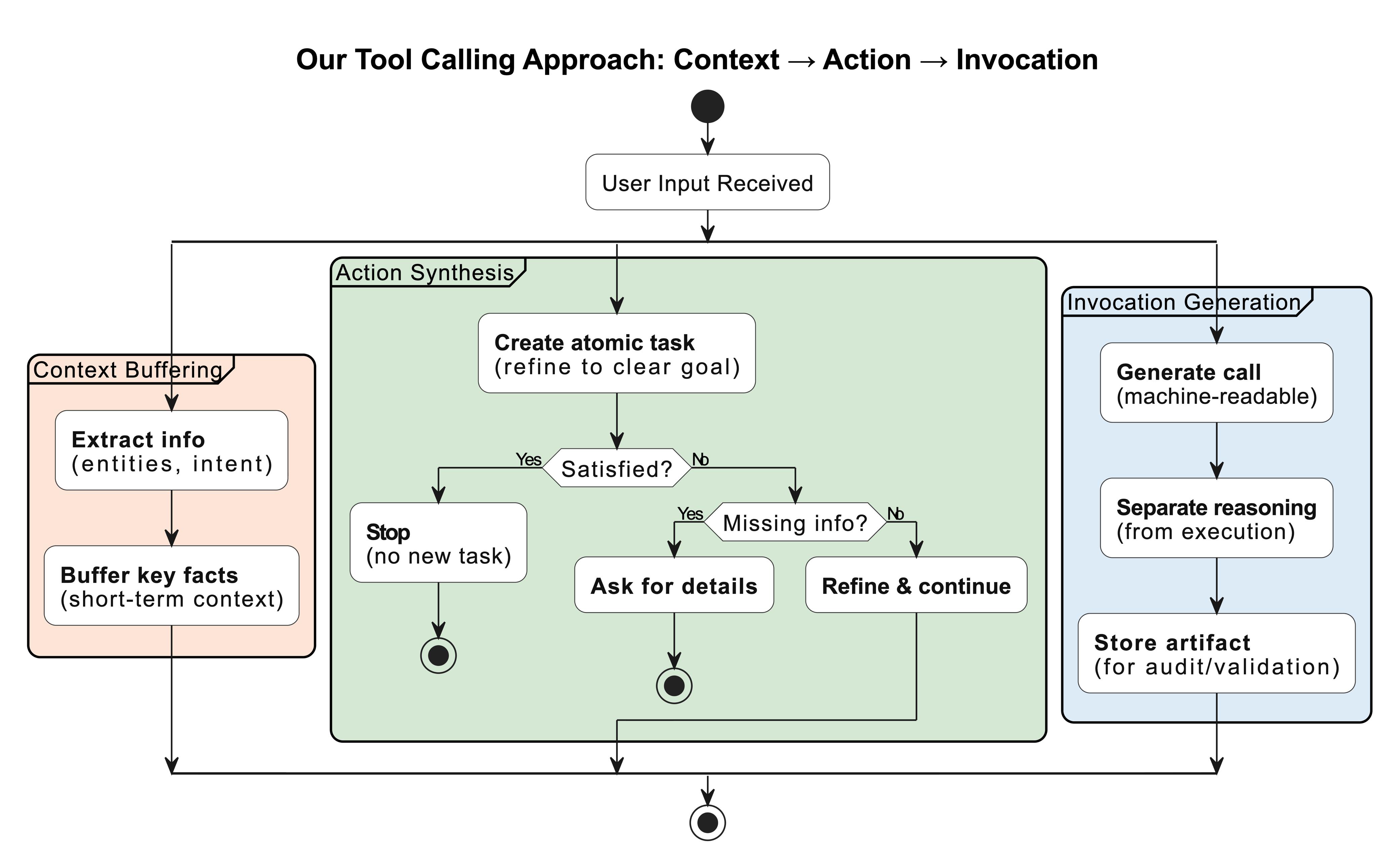}
\caption{The system processes conversational input through three stages: Context Buffering extracts and structures relevant information, Action Synthesis generates atomic, testable tasks while handling early termination and missing prerequisites, and Invocation Generation produces machine-interpretable actions with auditability, keeping reasoning and execution separated.}
\label{tc}
\end{figure}

To address these challenges, we introduce the \textbf{CoreThink Agentic Reasoner}, an updated version of the original CoreThink \cite{vaghasiya2025corethinksymbolicreasoninglayer}. Our reasoner augments large language models with a \textbf{symbolic reasoning layer} that explicitly decomposes complex, multi-step tasks, orchestrates specialized tools, and verifies intermediate results. This combination enables reliable multi-turn reasoning and out-of-distribution generalization while maintaining exceptional computational efficiency. CoreThink sets a new state-of-the-art across prominent agentic benchmarks, achieving performance improvements of 5--30\% over baseline models, and demonstrates robust performance on challenging OOD tasks.

This paper makes several key contributions. First, we present the CoreThink Agentic Reasoner, detailing its symbolic reasoning architecture and validating its superior performance on standard agentic benchmarks. Second, we introduce \textbf{MAVEN (Math \& Physics Adversarial Verification \& Evaluation Network)}, a novel adversarial, out-of-distribution dataset designed to stress-test reasoning systems on long-horizon math and physics problems. Our analysis shows that while many frontier models perform poor on MAVEN, CoreThink achieves state-of-the-art performance, demonstrating strong generalization and robustness. Third, we examine existing models on Agentic Tool Calling benchmarks and find that, while most struggle to perform reliably across diverse tasks, the CoreThink Agentic Reasoner demonstrates robust generalization. Finally, in the spirit of open science, we release MAVEN and our evaluation scripts to facilitate further research in multi-step, tool-augmented reasoning.

\section{Related Work}

\subsection{BFCL v3: Multi-Turn and Multi-Step Function Calling}

The Berkeley Function Calling Leaderboard (BFCL) v3 is a benchmark designed to assess large language models' (LLMs) ability to invoke external functions and APIs in response to user queries. It introduced multi-turn and multi-step evaluations, incorporating state-based tracking to assess models' performance in dynamic, real-world scenarios. This version expanded upon its predecessors by introducing augmented categories such as missing functions and long-context interactions, thereby providing a more comprehensive assessment framework.\cite{bfcl_v3}.

However, some critiques have emerged regarding the BFCL v3. For instance, the evaluation method relies heavily on Abstract Syntax Tree (AST)-based analysis, which may not fully capture the nuances of real-world function calling scenarios. Additionally, the benchmark's focus on specific programming languages and tools might limit its applicability across diverse domains and use cases.\cite{bfcl_critic}\cite{rabinovich2025robustnessagenticfunctioncalling}

\begin{figure}
\centering
\includegraphics[width=1.05\linewidth]{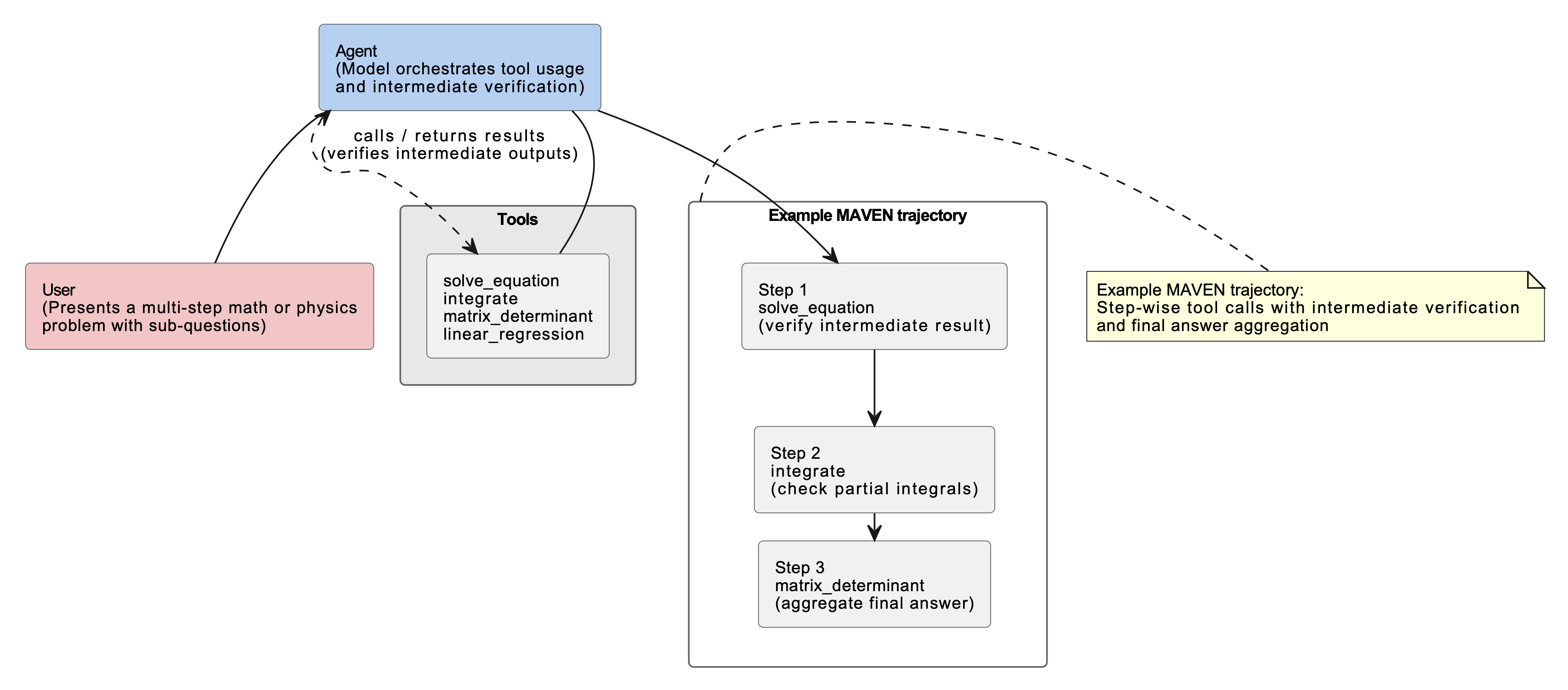}
\caption[Agent--tool orchestration for MAVEN]{Schematic of the MAVEN evaluation setup. A user supplies a multi-step math or physics problem; the Agent orchestrates calls to external tools (e.g., \texttt{solve\_equation}, \texttt{integrate}, \texttt{matrix\_determinant}, \texttt{linear\_regression}), verifies intermediate results at each step, and aggregates those results to produce the final solution. Right: an example MAVEN trajectory showing sequential, step-wise tool calls with intermediate verification and final answer aggregation.}
\label{setup}
\end{figure}

\subsection{TauBench: Tool-Agent-User Interaction}

$\tau$-Bench is a benchmark that emulates dynamic conversations between a user (simulated by language models) and a language agent provided with domain-specific API tools and policy guidelines. It aims to evaluate the agent's ability to solve complex tasks while adhering to domain-specific rules, which is crucial for deploying LLMs in real-world applications. The benchmark includes domains like retail and airline services, testing agents' interactions with simulated users and their adherence to domain-specific policies \cite{tau_bench}.

Despite its strengths, TauBench has faced criticism for its limited scope and potential lack of diversity in the scenarios it presents. Some researchers argue that the benchmark's focus on specific domains may not adequately test the generalization capabilities of LLMs across varied real-world applications. Moreover, the reliance on simulated users might not fully capture the complexities and unpredictability of human interactions, potentially affecting the benchmark's realism and relevance.\cite{survey}

\subsection{Tau2Bench: Dual-Control Benchmarking}

$\tau^2$-Bench extends the concept of $\tau$-Bench by introducing a dual-control framework, where both the agent and the user interact with a shared environment. This setup tests both agent coordination and communication, providing a more realistic evaluation of conversational AI agents. The benchmark includes domains such as mock, airline, retail, and telecom, each with specific policies and tools to assess the agent's performance \cite{tau2_bench}.

However, some critiques highlight the complexity of the dual-control framework, which may introduce challenges in evaluating agent performance consistently. The intricate interactions between the agent and user could lead to ambiguities in assessing the agent's capabilities, potentially affecting the reliability and validity of the benchmark results. Additionally, the benchmark's emphasis on specific domains might limit its applicability to broader contexts, necessitating further research to enhance its generalizability.\cite{survey}

\subsection{ACEBench: Comprehensive Tool Usage Evaluation}

ACEBench is a comprehensive benchmark designed to assess the function calling capabilities of LLMs. It addresses key limitations of existing evaluation systems, such as the lack of multi-turn dialogue assessments in real-world scenarios and the absence of fine-grained evaluations for parameter-type function calls. ACEBench categorizes evaluation scenarios into three types: Normal, Special, and Agent, enabling granular assessment across a wide spectrum of realistic tool-using scenarios \cite{acebench}.

Despite its comprehensive approach, ACEBench has been critiqued for its reliance on LLMs or real API executions for evaluation, which introduces significant overhead. This dependency could pose challenges in terms of scalability and efficiency, particularly when evaluating large models or conducting extensive assessments. Furthermore, the benchmark's focus on specific evaluation categories might not encompass all potential real-world scenarios, potentially limiting its applicability in diverse contexts.\cite{survey}

\section{Our Approach}

Our system translates conversational context into structured, verifiable actions, optionally producing executable invocations while minimizing unsafe side effects and maintaining auditability. The approach is organized around three main stages:

\begin{itemize}
\item \textbf{Context buffering:} The system first extracts and structures the key information from the conversation into a compact, short-lived buffer. This buffer preserves salient facts and any intermediate reasoning required for subsequent steps.

\item \textbf{Action synthesis:} Using the buffered context, the system generates an atomic, testable description of the task needed to fulfill the user’s request. It performs a bounded number of refinement attempts to ensure clarity and correctness while avoiding unnecessary iterations. If the system determines that no action is needed or the request is already satisfied, it terminates early.

\item \textbf{Invocation generation:} When all prerequisites are satisfied, the system produces a machine-interpretable invocation compatible with the available execution environment and adapters. By keeping reasoning and execution separate, the system reduces the risk of unintended side effects. A compact audit artifact is retained for validation, human review, and post-hoc analysis.
\end{itemize}

This staged design ensures that the system is both reliable and efficient, balancing careful reasoning with practical execution. Figure.\ref{tc}

\subsection{Evaluation Across multiple Benchmarks}
We evaluated CoreThink AI across multiple tool-calling benchmarks. The CoreThink system leverages a NeuroSymbolic reasoning layer on top of GPT-OSS-120b base model, combining large language model capabilities with symbolic reasoning to enhance task accuracy and tool orchestration.

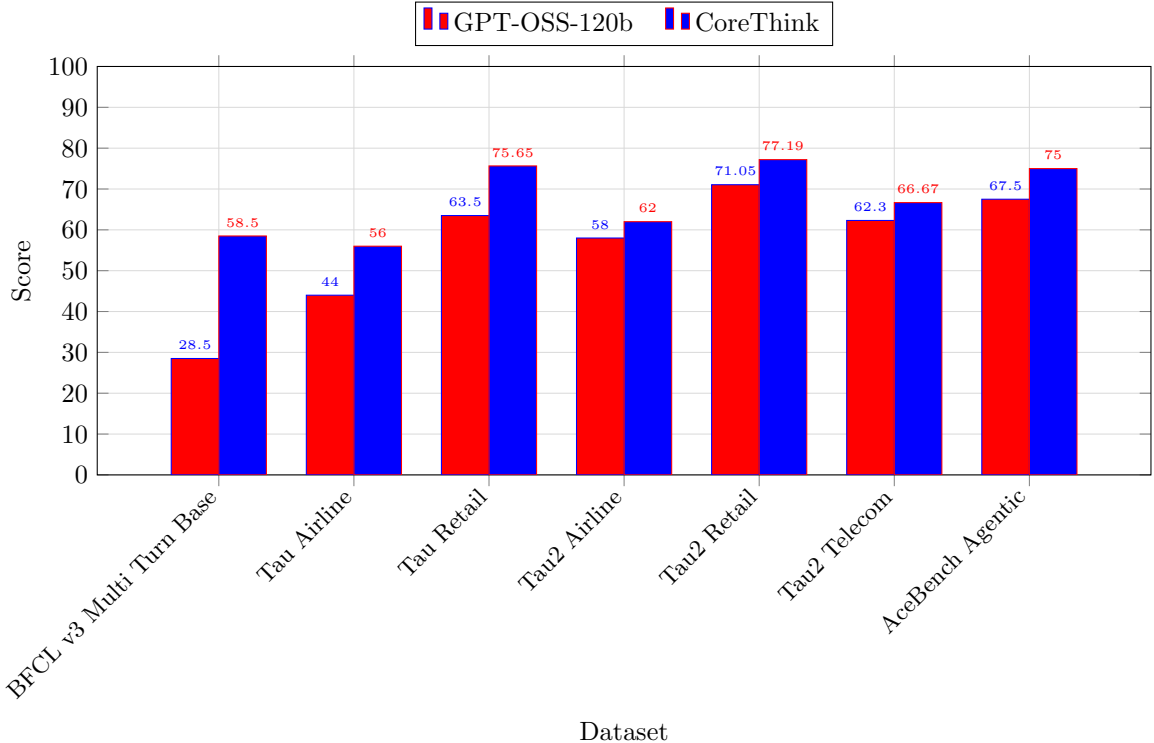
\begin{figure}[ht]
    \centering
\begin{tikzpicture}
\begin{axis}[
width=\textwidth,
height=0.45\textwidth,
ymin=0, ymax=100,
ylabel={Score},
xlabel={Dataset},
symbolic x coords={BFCL v3 Multi Turn Base, Tau Airline,Tau Retail,Tau2 Airline,Tau2 Retail,Tau2 Telecom,AceBench Agentic},
xtick=data,
xticklabel style={rotate=45,anchor=east,font=\small},
ytick={0,10,20,30,40,50,60,70,80,90,100},
grid=both,
grid style={line width=.1pt, draw=gray!10},
major grid style={line width=.2pt,draw=gray!30},
legend style={at={(0.5,1.05)}, anchor=south,legend columns=2, /tikz/every even column/.append style={column sep=1em}},
enlarge x limits=0.15,
bar width=18pt,
ybar=0pt,
nodes near coords={\tiny\pgfmathprintnumber\pgfplotspointmeta},
nodes near coords align={vertical}
]

\addplot+[ybar, fill=red] coordinates {
(BFCL v3 Multi Turn Base,28.50)
(Tau Airline,44.00)
(Tau Retail,63.50)
(Tau2 Airline,58.00)
(Tau2 Retail,71.05)
(Tau2 Telecom,62.30)
(AceBench Agentic,67.50)
};
\addlegendentry{GPT-OSS-120b}

\addplot+[ybar, fill=blue] coordinates {
(BFCL v3 Multi Turn Base,58.50)
(Tau Airline,56.00)
(Tau Retail,75.65)
(Tau2 Airline,62.00)
(Tau2 Retail,77.19)
(Tau2 Telecom,66.67)
(AceBench Agentic,75.00)
};
\addlegendentry{CoreThink}

\end{axis}
\end{tikzpicture}
    \caption{Comparison of CoreThink AI and GPT-OSS-120b scores across tool-calling benchmarks. CoreThink consistently outperforms the base LLM, demonstrating the benefits of its NeuroSymbolic reasoning layer.}
    \label{fig:CoreThink_vs_gptoss}
\end{figure}

\begin{table}[ht]
\centering
\caption{Comparison of Model Scores Across Domains}
\label{tab:model-scores-expanded}
\small
\begin{adjustbox}{width=\textwidth}
\begin{tabular}{|l|l|*{8}{c|}}
\hline
\textbf{Model} & \textbf{Domain} & 
\rotatebox{20}{\textbf{CoreThink}} & 
\rotatebox{20}{\textbf{GPT-5}} &
\rotatebox{20}{\textbf{o4-mini}} &
\rotatebox{20}{\textbf{o3}} &
\rotatebox{20}{\textbf{Kimi-K2}} &
\rotatebox{20}{\textbf{Deepseek-V3.1}} &
\rotatebox{20}{\textbf{Qwen3-Thinking-235B}} &
\rotatebox{20}{\textbf{Gemini-2.5-pro}} \\
\hline
Tau & Airline & \textbf{56.00} & 44 & 46.00 & 52.00 & 51.20 & 40.00 & 46.00 & 44.00 \\
Tau & Retail  & 75.65 & \textbf{78.3} & 70.40 & 70.40 & 73.90 & 66.10 & 67.80 & 68.70 \\
\hline
Tau2 & Airline & \textbf{62.00} & \textbf{62.00} & 46.00 & 52.00 & 51.20 & 40.00 & 46.00 & 44.00 \\
Tau2 & Retail  & 77.19 & \textbf{81.1} & 70.40 & 70.40 & 73.90 & 66.10 & 67.80 & 68.70 \\
Tau2 & Telecom & 66.67 & \textbf{96.7} & 46.50 & 58.20 & 65.80 & 38.50 & 45.60 & 27.20 \\
\hline
BFCL v3 & Multi Turn Base & 58.50 & 33.5 & 53.00 & 44.00 & \textbf{60.50} & 44.00 & 53.50 & 35.00 \\
\hline
AceBench & Agentic & \textbf{75.00} & 32.5 & 60.00 & 63.30 & 65.00 & 40.80 & 39.10 & 63.40 \\
\hline
\textbf{Overall} & - & \textbf{67.28} & 61.15 & 56.04 & 58.61 & 63.07 & 47.93 & 52.26 & 50.14 \\
\hline
\end{tabular}
\end{adjustbox}
\end{table}
Across all datasets, CoreThink consistently outperforms GPT-OSS-120B, demonstrating significant gains ranging from 5\% to 30\%, depending on the domain. The most notable improvements are observed in multi-turn and agentic reasoning tasks, where symbolic verification and tool orchestration play a critical role. Figure.\ref{fig:CoreThink_vs_gptoss}
Most of the scores in the table above are taken from Runnan Fang et al. \cite{agentscaler}, and, except for AceBench, all experiments were run in the Function Calling (FC) format.
\subsection{Performance Gains through Neuro-Symbolic Reasoning}

The CoreThink NeuroSymbolic layer enhances the GPT-OSS-120b base by:
\begin{enumerate}
\item \textbf{Structured Reasoning:} Verifies intermediate steps when invoking external tools to prevent error propagation.
\item \textbf{Adaptive Tool Orchestration:} Dynamically selects and sequences tool calls based on problem structure, improving correctness and efficiency.
\item \textbf{Cross-Domain Consistency:} Ensures logical coherence across multiple domains, contributing to higher reliability in complex tasks.
\end{enumerate}

Overall, the integration of symbolic reasoning on top of a high-capacity LLM allows CoreThink to achieve \textbf{state-of-the-art performance}, surpassing GPT-OSS-120b across all evaluated benchmarks while maintaining strong interpretability of its reasoning steps.

\section{MAVEN Dataset}
\label{sec:maven}
MAVEN (Math \& Physics Adversarial Verification \& Evaluation Network) is an evaluation ecosystem explicitly designed to measure the ability of tool-using agents to perform extended, verifiable scientific problem solving. The benchmark concentrates on three interlocking competencies: reliable orchestration of multiple specialised computational tools, disciplined preservation and inspection of intermediate state, and the explicit verification of intermediate results that together produce reproducible final outcomes. Unlike short-question corpora that can encourage rote pattern matching or single-step retrieval, MAVEN targets sustained chains of reasoning that are representative of real scientific workflows where diagnostic awareness and provenance are essential.

\begin{figure}
\caption{Example (schematic)} The following is a compact representation of a canonical MAVEN instance:
\begin{lstlisting}
{
  "id": "MAVEN-0001",
  "statement": "A particle moves along x(t)=A t^3 - B t^2 + C t. Given A,B,C and initial conditions, find the time of local maxima of kinetic energy and compute the kinetic energy at that time.",
  "sub_questions": [
    "1. Compute v(t)=dx/dt.",
    "2. Compute K(t)=0.5 m v(t)^2.",
    "3. Solve dK/dt = 0 for t (identify candidate times).",
    "4. Verify second derivative condition for maxima.",
    "5. Evaluate K(t) at the verified time(s)."
  ],
  "required_tools": ["symbolic_diff","algebra_solver","numeric_evaluator"],
  "reference_solution": { ... }
}
\end{lstlisting}
\label{fig:maven-schematic}
\end{figure}

\subsection{Motivation}
The central motivation behind MAVEN is to provide a rigorous benchmark for evaluating agents on multi-step reasoning tasks, particularly under out-of-distribution (OOD) conditions. In real-world scientific and engineering scenarios, it is not enough for an agent to produce a correct final answer; the sequence of intermediate decisions, the reasoning process, and the handling of edge-case situations are equally critical. MAVEN is designed to stress-test these capabilities by constructing tasks that require agents to integrate symbolic, numeric, and tool-augmented reasoning across extended solution traces. By emphasizing OOD generalization, persistent representation management, and explicit verification at multiple checkpoints, MAVEN measures not only what an agent outputs, but how reliably and systematically it arrives at that output. Evaluating performance on OOD tasks also allows us to detect overfitting, since comparing results on training versus unseen tasks highlights whether the agent has learned generalizable reasoning strategies or is merely memorizing patterns.
\subsection{Dataset composition and parametric instantiation}
MAVEN's core corpus contains one hundred canonical problem templates drawn from calculus, algebra, linear algebra, classical mechanics, thermodynamics, electromagnetism and diverse branches of applied mathematics. Each canonical problem is intentionally parameterised so that concrete instantiations vary in numerical regime, algebraic form and verification requirements. A single MAVEN template therefore produces many distinct test cases: small changes in parameters can induce ill-conditioning, create near-degenerate stationary points, or produce multiple algebraic branches requiring disambiguation. This parametric design deliberately raises the bar for generalisation: success on MAVEN requires robust strategies for tool selection, conditioning-aware numerical computation, and explicit verification that generalise across instantiations rather than brittle memorisation of prompt–answer pairs.

\subsection{Model Context Protocol (MCP) and persistent state}
A defining component of the MAVEN release is the Model Context Protocol (MCP), a protocol and reference implementation that formalises how agents persist, query and reason about intermediate results. MCP treats intermediate artifacts as first-class objects: symbolic expressions, numerics with units, solver diagnostics and provenance metadata are each stored under explicit step identifiers and made available for later retrieval. This persistence model supports experimental questions that are otherwise difficult to study in isolation: how often should agents recompute versus reuse persisted results, which forms of intermediate representation improve downstream stability, and how does explicit provenance aid automated verification? By including a dockerised MCP server and client examples, MAVEN enables reproducible evaluation of these questions in a way that makes the experimental assumptions explicit and inspectable.

\subsection{Construction and validation pipeline}
Problems in MAVEN were produced by a multi-stage pipeline that mixes human expertise and automated validation. Domain specialists drafted seeds that require multi-concept reasoning and identified canonical tool-paths that typify safe solution strategies. These seeds were then augmented adversarially by injecting distractor terms, creating parameter regimes that stress numerical stability, and permuting algebraic forms to produce alternative but valid solution branches. Canonical traces—sequences of tool calls together with expected intermediate outputs—were executed in the MCP sandbox to generate ground-truth artifacts. Automated perturbation tests followed, running canonical paths under variations of solver tolerances, input perturbations and ordering permutations to ensure the reference answers and verification checkpoints are stable under plausible variations. Finally, independent human reviewers audited both the canonical traces and the validation harness to ensure that the recorded traces capture meaningful, domain-relevant reasoning steps rather than artefacts of particular solver implementations.

\begin{figure}
\caption{Minimal MCP interaction example (schematic)}
\begin{lstlisting}
# Step 1: call a tool and persist its output in the MCP context
POST /mcp/call
Body: {
  "problem_id": "MAVEN-0001",
  "step_id": "step-01",
  "tool_id": "symbolic_diff",
  "input": { "expr": "A*t^3 - B*t^2 + C*t", "wrt": "t" },
  "persist": true
}
Response: {
  "ok": true,
  "result_id": "MAVEN-0001-step-01-result",
  "output": { "expr": "3*A*t^2 - 2*B*t + C" },
  "diagnostics": { "type": "symbolic", "simplified": true }
}

# Step 2: query MCP for step-01 outputs and use them as input
POST /mcp-server/mcp
Body: {
  "problem_id": "MAVEN-0001",
  "query": { "from_step": "step-01", "fields": ["output.expr"] }
}
Response: {
  "ok": true,
  "matches": [
    { "result_id": "MAVEN-0001-step-01-result", "output": { "expr": "3*A*t^2 - 2*B*t + C" } }
  ]
}
\end{lstlisting}
\label{mcp}
\end{figure}

\subsection{Annotation, diagnostics and failure-mode disclosure}
Every MAVEN instance is distributed with a rich annotation bundle. Step-level traces record the tool invoked, the exact input payload, expected outputs annotated with acceptable equivalence classes and recommended numeric tolerances. Crucially, each persisted artifact includes diagnostic metadata such as solver status flags, convergence metrics, condition numbers and simplification provenance. We treat failure modes as first-class information: known pitfalls (for example, divisions by small numbers, ambiguous branch cuts, or near-singular matrices) are documented alongside recommended checks. This emphasis on diagnostics and explicit failure-mode disclosure serves two purposes. Practically, it allows evaluators to distinguish between a correct-looking final answer produced by reckless numerics and a robustly verified result. Scientifically, it provides a substrate for studying how explicit diagnostic information affects agent behaviour and error-correction strategies.

\subsection{Evaluation protocol and metrics}
MAVEN evaluates agents along multiple, complementary axes that reflect the multifaceted nature of reliable computation. Sub-question accuracy measures correctness at the granularity of individual verification checkpoints and intermediate deliverables. Tool selection accuracy captures whether the chosen primitive is appropriate for the subtask at hand. Trace fidelity quantifies alignment between the agent's MCP trace and the canonical reference while tolerating legitimate variations in ordering and representation. Verification score measures whether explicit checks—units consistency, second-derivative tests for extrema, or solver residual-based acceptance criteria—were both executed and correctly interpreted. Final-answer correctness employs symbolic-equivalence checks and parameterised numeric tolerances. 
\subsection{On generalisation and the risk of narrow overfitting}
Concerns about overfitting are both reasonable and explicitly addressed in MAVEN's conception and release. The dataset designers resisted the tempting optimisation of producing superficially diverse problems that nonetheless share narrow solution templates. Instead, MAVEN emphasises parametric richness, multiple valid tool-paths and adversarial perturbations that change the underlying numerical or algebraic character of instances. Because canonical traces are seldom unique, an agent that simply memorises a single path will perform poorly under instance perturbation and on metrics that reward explicit verification and provenance. Moreover, the benchmark authors intentionally validated that trivial fixes to a single agent do not yield broad improvements across the full suite of MAVEN metrics; substantive gains require improvements in verification strategy, conditioning-aware numerics, or provenance-aware state management. To foster transparency, the full MCP traces, problem generators and validation harness are released so the community can audit instance distributions, propose alternative canonical traces and replicate experiments.
\section{Evaluation}
\label{sec:evaluation}

We evaluate systems on MAVEN using a purpose-built evaluation harness to measure tool-centric problem solving on long, multi-step mathematics and physics problems. The MAVEN combines strict execution protocols, an expert automated judge (GPT-4.1), and advanced rule-violation handling to produce reproducible, trace-aware assessments focused on tool orchestration, numerical correctness, and methodological rigor. GPT-4.1 was also used as a judge in HealthBench and we continue to use it. \cite{health}

\subsection{Testing protocol and strict rules}
To ensure fairness and comparability across models, every test is executed under the following high-level constraints:
\begin{enumerate}
  \item \textbf{Tools-only constraint:} Models are not permitted to perform manual arithmetic or ad-hoc symbolic manipulation outside the sanctioned tool set; all computations must be performed via explicit tool calls exposed by the MCP.
  \item \textbf{Single-call per response:} Each model response may contain at most one tool invocation. This constraint enforces sequential, verifiable decomposition of multi-step problems and simplifies trace analysis.
  \item \textbf{Completion signal:} Agents must return a canonical completion marker (``\texttt{PROBLEM\_COMPLETED}'') when they have reached a final answer; the marker distinguishes partial traces from completed solutions.
  \item \textbf{Deterministic tool environment:} The MCP tool implementations are deterministic and versioned so that identical inputs produce identical outputs, facilitating reproducibility and trace comparison.
\end{enumerate}

Violations of the single-call rule or the tools-only constraint are handled systematically (Section~\ref{sec:trace-recon}), enabling fair scoring even when agents do not strictly follow the protocol.

\subsection{Rule violation handling and trace reconstruction}
\label{sec:trace-recon}
A novel feature of the MAVEN is its \emph{execution-trace reconstruction} pipeline. When models violate the single-call or tools-only constraints (for example, by embedding multiple implicit computations in one response), the harness attempts to reconstruct a plausible tool-by-tool execution trace from the model output and any available intermediate artifacts persisted to the MCP. Reconstructed traces are then scored under the same rubric; a separate diagnostic flag is recorded to indicate that reconstruction was necessary. This approach mitigates unfairly punitive outcomes for otherwise capable agents while preserving incentives to adhere to the canonical interaction protocol.

\subsection{Experimental setup, concurrency, and reproducibility}
Evaluation supports configurable experimental regimes (e.g., selective model lists, parallel execution, and resumable runs) while maintaining reproducibility guarantees:
\begin{itemize}
  \item \textbf{Controlled tool environment:} all tools exposed by the MCP are versioned and instrumented to record provenance (tool id, version, inputs, outputs, diagnostics).
  \item \textbf{Deterministic execution:} random seeds and timeouts are controlled when tools are stochastic or iterative; default time budgets prevent unbounded computation per question.
  \item \textbf{Resumable runs:} long experiments can be paused and resumed without loss of previously computed traces or scores, enabling robust large-scale evaluation.
  \item \textbf{Calibration and validation:} we calibrate judge prompts and scoring thresholds using a held-out set of calibration problems and perform periodic human audits (double-judging a random subset) to validate automated judgments.
\end{itemize}

\subsection{Outputs and artifacts}
For each evaluated model the MAVEN Tester produces:
\begin{itemize}
  \item Per-response scored records with dimension-wise breakdown (tools, correctness, approach), judge critique, and diagnostic tags.
  \item Aggregate summaries: per-model mean and median scores, accuracy rates, completion rates, and error-mode histograms.
  \item Canonical and reconstructed MCP traces for trace-level analysis and reproducibility.
  \item Standardized result tables (CSV) suitable for downstream statistical analysis.
\end{itemize}

\subsection{Leaderboard and representative results}
Table~\ref{tab:leaderboard} reports representative results obtained from a set of vetted models evaluated on MAVEN. These results illustrate the sensitivity of MAVEN to tool orchestration behavior and the discriminative power of the tool-centric rubric. Pricing via OpenRouter as of October 15, 2025.

\begin{table}[ht]
\centering
\caption{Representative model results on MAVEN}
\label{tab:leaderboard}
\renewcommand{\arraystretch}{1.2}
\resizebox{\textwidth}{!}{
\begin{tabular}{|l|c|c|c|c|c|c|}
\hline
\textbf{Model} & \textbf{Accuracy (\%)} & \textbf{Partial Scoring (/100)} & \textbf{Tool Usage (/70)} & \textbf{Correctness (/20)} & \textbf{Approach (/10)} & \textbf{Approx Price}\\
\hline
\textbf{CoreThink/openai/gpt-oss-120b} & \textbf{71.0} & \textbf{95.1} & \textbf{67.4} & \textbf{18.2} & \textbf{9.5} & \$1.5\\
\hline
Anthropic/Claude-Sonnet-4.5 & \textbf{70.0} & \textbf{94.2} & \textbf{67} & \textbf{17.7} & \textbf{9.5} & \$15\\
\hline
Moonshotai/kimi-k2-0905\cite{kimi} & 57.0 & 92.4 & 65.5 & 17.7 & 9.3 & \$ 5\\
\hline
X-AI/grok-4\cite{xai2025grok4} & 55.0 & 88.2 & 63 & 16.8 & 8.4 & \$ 15\\
\hline
OpenAI/gpt-oss-120b\cite{oss} & 48.0 & 88.9 & 62.6 & 17.5 & 8.7 & \$0.9\\
\hline
Z-AI/glm-4.5\cite{glm45} & 43.0 & 88.0 & 62.3 & 16.9 & 8.7 & \$3\\
\hline
OpenAI/o4-mini\cite{o4} & 38.0 & 76.7 & 53.6 & 15.7 & 7.5 & \$9\\
\hline
OpenAI/gpt-5\cite{gpt5} & 32.0 & 61.1 & 43.3 & 12.6 & 5.2 & \$15\\
\hline
\end{tabular}
}
\end{table}

\subsection{Analysis and observed failure modes}
Across evaluated systems, we observe several recurrent failure modes:
\begin{itemize}
  \item \textbf{Tool-selection errors:} choosing a numerically unstable solver or an ill-suited symbolic routine leads to low tool-usage scores even if the final numerical answer is sometimes recoverable.
  \item \textbf{Missing verification:} agents frequently omit critical verification steps (e.g., sign checks, second-derivative tests), which reduces the approach score and increases susceptibility to subtle correctness errors.
  \item \textbf{Rule violations:} higher-performing, unconstrained models occasionally violate the single-call rule; reconstruction mitigates but does not fully eliminate the penalty associated with protocol non-compliance.
  \item \textbf{Numerical instability:} certain adversarial parameter regimes in MAVEN intentionally expose conditioning issues; robust agents must both detect and adapt to such warnings to score highly.
\end{itemize}

\section{Generalization using CoreThink Reasoning Layer}
We integrated the CoreThink Agentic Reasoning Layer into these models. Experimental results demonstrate a marked improvement in performance across both benchmarks, suggesting that the CoreThink layer effectively augments the models’ ability to decompose and execute complex reasoning tasks. These findings provide strong evidence for the generalizability of the CoreThink framework, indicating that it can systematically enhance multi-step reasoning across diverse large language models.

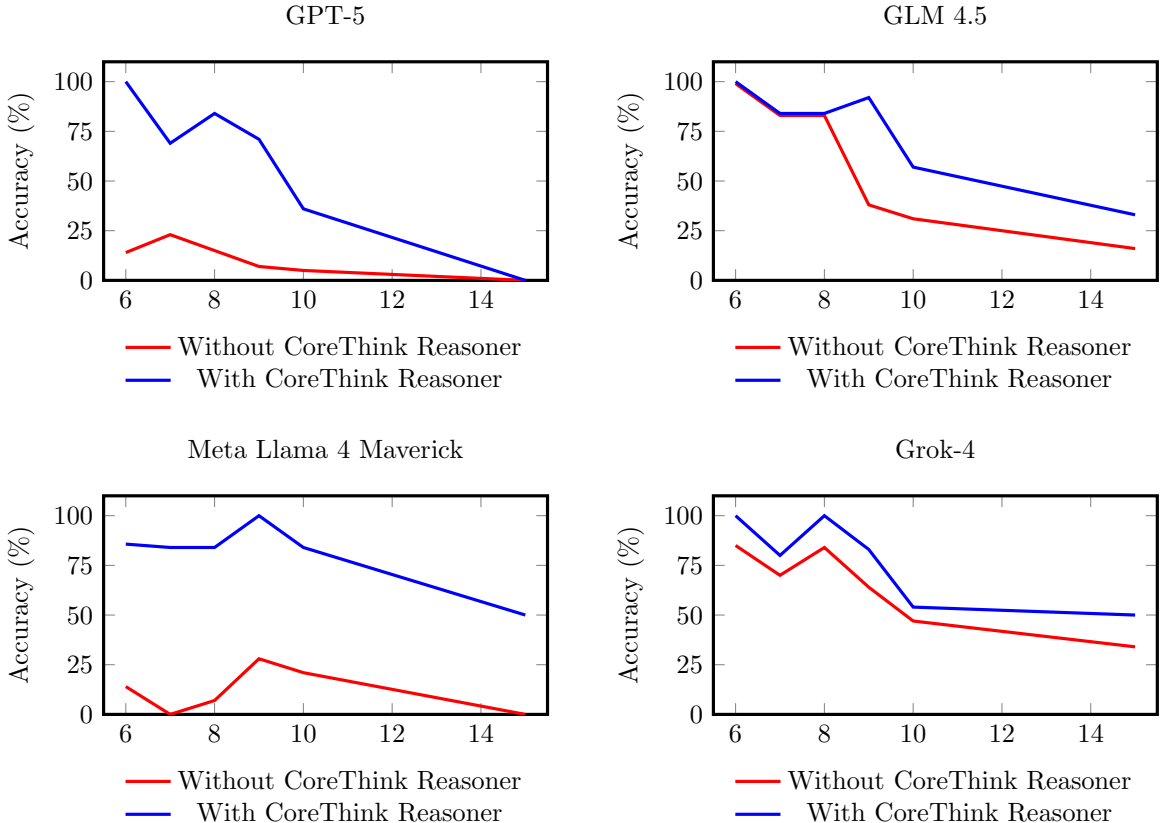
\begin{figure}[ht]
    \centering
    \begin{subfigure}{0.48\textwidth}
        \centering
        \begin{tikzpicture}
            \begin{axis}[
                width=\textwidth,
                height=0.6\textwidth,
                xlabel={Steps},
                ylabel={Accuracy (\%)},
                title={GPT-5},
                title style={yshift=1ex},
                ymin=0, ymax=110,
                xmin=5.5, xmax=15.5,
                ytick={0,25,50,75,100},
                grid=none,
                legend style={at={(0.5,-0.2)}, anchor=north, draw=none},
                line width=1.2pt,
            ]
                \addplot[color=red] coordinates {
                    (6,14)
                    (7,23)
                    (8,15)
                    (9,7)
                    (10,5)
                    (15,0)
                };
                \addlegendentry{Without CoreThink Reasoner}

                \addplot[color=blue] coordinates {
                    (6,100)
                    (7,69)
                    (8,84)
                    (9,71)
                    (10,36)
                    (15,0)
                };
                \addlegendentry{With CoreThink Reasoner}
            \end{axis}
        \end{tikzpicture}
    \end{subfigure}
    \hfill
    \begin{subfigure}{0.48\textwidth}
        \centering
        \begin{tikzpicture}
            \begin{axis}[
                width=\textwidth,
                height=0.6\textwidth,
                xlabel={Steps},
                ylabel={Accuracy (\%)},
               title={GLM 4.5},
                title style={yshift=1ex},
                ymin=0, ymax=110,
                xmin=5.5, xmax=15.5,
                ytick={0,25,50,75,100},
                grid=none,
                legend style={at={(0.5,-0.2)}, anchor=north, draw=none},
                line width=1.2pt,
            ]
                \addplot[color=red] coordinates {
                    (6,99)
                    (7,83)
                    (8,83)
                    (9,38)
                    (10,31)
                    (15,16)
                };
                \addlegendentry{Without CoreThink Reasoner}

                \addplot[color=blue] coordinates {
                    (6,100)
                    (7,84)
                    (8,84)
                    (9,92)
                    (10,57)
                    (15,33)
                };
                \addlegendentry{With CoreThink Reasoner}
            \end{axis}
        \end{tikzpicture}
    \end{subfigure}

    \vskip1em
    \begin{subfigure}{0.48\textwidth}
        \centering
        \begin{tikzpicture}
            \begin{axis}[
                width=\textwidth,
                height=0.6\textwidth,
                xlabel={Steps},
                ylabel={Accuracy (\%)},
                title={Meta Llama 4 Maverick},
                title style={yshift=1ex},
                ymin=0, ymax=110,
                xmin=5.5, xmax=15.5,
                ytick={0,25,50,75,100},
                grid=none,
                legend style={at={(0.5,-0.2)}, anchor=north, draw=none},
                line width=1.2pt,
            ]
                \addplot[color=red] coordinates {
                    (6,14)
                    (7,0.0)
                    (8,7)
                    (9,28)
                    (10,21)
                    (15,0)
                };
                \addlegendentry{Without CoreThink Reasoner}

                \addplot[color=blue] coordinates {
                    (6,85.7)
                    (7,84)
                    (8,84)
                    (9,100)
                    (10,84)
                    (15,50)
                };
                \addlegendentry{With CoreThink Reasoner}
            \end{axis}
        \end{tikzpicture}
    \end{subfigure}
    \hfill
    \begin{subfigure}{0.48\textwidth}
        \centering
        \begin{tikzpicture}
            \begin{axis}[
                width=\textwidth,
                height=0.6\textwidth,
                xlabel={Steps},
                ylabel={Accuracy (\%)},
                title={Grok-4},
                title style={yshift=1ex},
                ymin=0, ymax=110,
                xmin=5.5, xmax=15.5,
                ytick={0,25,50,75,100},
                grid=none,
                legend style={at={(0.5,-0.2)}, anchor=north, draw=none},
                line width=1.2pt,
            ]
                \addplot[color=red] coordinates {
                    (6,85)
                    (7,70)
                    (8,84)
                    (9,64)
                    (10,47)
                    (15,34)
                };
                \addlegendentry{Without CoreThink Reasoner}

                \addplot[color=blue] coordinates {
                    (6,100)
                    (7,80)
                    (8,100)
                    (9,83)
                    (10,54.0)
                    (15,50)
                };
                \addlegendentry{With CoreThink Reasoner}
            \end{axis}
        \end{tikzpicture}
    \end{subfigure}

    \caption{LLM comparison plots on MAVEN as the minimum number of steps required to solve increases}
    \label{fig:grid}
\end{figure}

\begin{table}[h!]
\centering
\begin{tabular}{|l|c|}
\hline
\multicolumn{2}{|c|}{\textbf{BFCL Multi Turn Base}} \\
\hline
CoreThink/Openai/gpt-5 & 51.5 \\
Openai/gpt-5 & 33.5 \\
\hline
CoreThink/Meta/Llama-4-Maverick & 46 \\
Meta/Llama-4-Maverick & 23.5 \\
\hline
\multicolumn{2}{|c|}{\textbf{MAVEN}} \\
\hline
CoreThink/Openai/gpt-5 & 66 \\
Openai/gpt-5 & 32 \\
\hline
CoreThink/Zai/GLM-4.5 & 59 \\
Zai/GLM-4.5 & 43 \\
\hline
CoreThink/Xai/Grok-4 & 69 \\
Xai/Grok-4 & 55 \\
\hline
CoreThink/Meta/Llama-4-Maverick & 54 \\
Meta/Llama-4-Maverick & 6 \\
\hline
\end{tabular}
\caption{Performance Improvement in the models using CoreThink Agentic Reasoner}
\end{table}
We also tried to find the accuracy of the models with and without using CoreThink Reasoning layer as the problem complexity increase Figure.\ref{fig:grid}. We see as the number of minimun required steps to solve the problem increases, accuracy of all the models reduces but with CoreThink Reasoner, the drop is less.
\section{Discussion}

The results from our evaluations highlight several tentative observations about the current landscape of agentic reasoning systems. While large-scale language models demonstrate impressive raw capabilities, their reasoning proficiency appears potentially brittle under dynamically verifiable, multi-step conditions such as those presented in MAVEN. Notably, many models, including GLM4.5, GPT-5, Kimi-K2, and Grok-4, achieve high scores on conventional benchmarks like Tau, yet show marked performance degradation on MAVEN. The partial score of many of these models is excellent but solving the complete question remains a challenge. Some models do good on some benchmarks, while others on different benchmarks. This suggests that benchmark saturation in existing datasets may reflect overfitting rather than genuine reasoning in some LLMs. But on the other hand, there are models which do well on all the benchmarks and can be said have achieved true generalization like Claude-Sonnet-4.5

In contrast, the \textbf{CoreThink Agentic Reasoner} tends to exhibit stronger and more consistent generalization across all the benchmarks. Its integration of a \textbf{NeuroSymbolic reasoning layer} atop a large-scale LLM backbone appears to contribute to improved interpretability and reliability by enabling intermediate verification, structured context persistence, and adaptive tool orchestration. These mechanisms may mitigate error propagation, which often challenges purely neural systems, especially in long-horizon reasoning tasks.

The \textbf{MAVEN dataset} is particularly informative in revealing these differences. Its design---emphasizing multi-step decomposition, adversarial edge cases, and explicit verification requirements---appears sensitive to subtle reasoning failures. Models that perform well on traditional leaderboards frequently struggle with tool selection, numerical stability, or rule compliance under MAVEN evaluations. This underscores MAVEN’s potential value as a more rigorous standard for assessing scientific reasoning agents.

Another noteworthy consideration is \textbf{efficiency and accessibility}. CoreThink achieves competitive results while operating at approximately one-tenth the computational cost of leading models. This efficiency derives in part from its reliance on smaller, open-source models, which significantly reduce hardware and energy requirements. Such characteristics enhance accessibility, facilitate reproducible experimentation on modest hardware, and allow for modular integration into domain-specific research pipelines.

Looking forward, these findings emphasize the importance of \textbf{dynamic, process-aware evaluation frameworks} that go beyond static answer checking. Benchmarks like MAVEN encourage the community to consider reasoning \emph{process fidelity} alongside final correctness. The open-sourcing of MAVEN and accompanying evaluation scripts is intended to foster a collaborative ecosystem where reasoning agents can be tested under transparent, reproducible, and adversarially challenging conditions.

Ultimately, the \textbf{CoreThink Agentic Reasoner} represents a step toward resilient, interpretable, and efficient agentic systems---ones capable not only of solving problems but also of \emph{understanding} and \emph{verifying} their reasoning paths. We hope this work will stimulate further exploration of dynamic scientific evaluation, bridging the gap between static benchmarks and adaptive, real-world reasoning agents.

\section*{Availability}
The source code for MAVEN is available at:  
\href{https://github.com/CoreThink-AI/adversarial-function-calling-benchmark}{https://github.com/CoreThink-AI/maven-benchmark}

\section*{Acknowledgment}

The authors would like to thank Ram Shanmugam and Chandra Khatri for their helpful discussions and contributions to this work. This research was supported by CoreThink AI.

\printbibliography

\end{document}